\documentclass{article}

\usepackage[a4paper, total={6in, 8in}]{geometry}
\usepackage[utf8]{inputenc}
\usepackage[T1]{fontenc}
\usepackage{hyperref}
\usepackage{amsfonts}
\usepackage{parskip}
\usepackage{bm}
\usepackage{amsmath}
\usepackage{verbatim}
\usepackage{listings}
\usepackage{subcaption}
\usepackage{graphicx}
\usepackage{amssymb}
\usepackage{empheq}
\usepackage{amsthm}
\usepackage{url}
\usepackage{authblk}

\usepackage{xcolor}
\hypersetup{
	colorlinks,
	linkcolor={red!50!black},
	citecolor={blue!50!black},
	urlcolor={blue!80!black}
}

\title{N-ODE Transformer: A Depth-Adaptive Variant of the Transformer Using Neural Ordinary Differential Equations}
\author[1]{Aaron Baier-Reinio}
\author[1]{Hans De Sterck}
\affil[1]{Department of Applied Mathematics, University of Waterloo, Waterloo N2L 3G1, Canada}

\begin{document}
	
\maketitle

\begin{abstract}
\noindent
We use neural ordinary differential equations to formulate a variant of the Transformer that is depth-adaptive
in the sense that an input-dependent number of time steps is taken by the ordinary differential equation solver. 
Our goal in proposing the N-ODE Transformer is to investigate whether its depth-adaptivity may aid in
overcoming some specific known theoretical limitations of the Transformer in handling nonlocal effects.
Specifically, we consider the simple problem of determining the parity of a binary sequence,
for which the standard Transformer has known limitations that can only be overcome by using a
sufficiently large number of layers or attention heads. We find, however, that the depth-adaptivity of the
N-ODE Transformer does not provide a remedy for the inherently nonlocal nature of the parity problem,
and provide explanations for why this is so.
Next, we pursue regularization of the N-ODE Transformer by penalizing the arclength of the ODE trajectories,
but find that this fails to improve the accuracy or efficiency of the N-ODE Transformer on the challenging parity
problem. We suggest future avenues of research for modifications and extensions of the N-ODE Transformer
that may lead to improved accuracy and efficiency for sequence modelling tasks such as neural machine translation.
\end{abstract}

\section{Introduction}

Introduced in 2017, the Transformer \cite{transformer} is an attention-based neural network commonly used for sequence modelling. In many sequence modelling tasks, Transformers are preferred over other kinds of neural networks such as recurrent neural networks (RNNs), because they are parallelizable and tend to be faster to train. However, the computational efficiency of Transformers has been put under scrutiny in recent years, and many variants of the Transformer that are better suited for processing very long sequences have been proposed. Many of the proposed variants of the Transformer fall under two categories:

\begin{itemize}
	\item Improving the $O(L^2)$ computational complexity of multi-headed attention (\cite{sparse_transformer}, \cite{reformer}, \cite{longformer}), where $L$ is the length of the input sequence.
	\item Designing a Transformer that adaptively changes its depth based on the input sequence (\cite{universal_transformer}, \cite{depth_adaptive_transformer}).
\end{itemize}

In this paper we are interested in the latter category. Namely, we propose to use neural ordinary differential equations \cite{nodes} (N-ODEs) in order to obtain a Transformer architecture that is, in a specific sense, depth-adaptive. \par

The paper is organized as follows. In Section \ref{sec:tf} we introduce the vanilla Transformer architecture and we discuss why a depth-adaptive Transformer is of interest. In Section \ref{sec:node_tf} we propose a new depth-adaptive variant of the Transformer using N-ODEs. In Section \ref{sec:rhs} we empirically compare different configurations of the N-ODE Transformer in order to decide upon which configurations we should use going forward. Section \ref{sec:vanilla_vs_N-ODE} empirically compares the vanilla Transformer to the N-ODE Transformer in terms of accuracy and training time. Section \ref{sec:reg} studies an approach for regularizing the N-ODE trajectories, in order to achieve a higher accuracy and take fewer time steps. In Section \ref{sec:disc} we discuss our findings and comment on future work. \par

\section{Relevant Aspects of the Vanilla Transformer} \label{sec:tf}

As is discussed in $\cite{transformer}$, the Transformer consists of two main components: the encoder and the decoder. For reasons mentioned in Section \ref{sec:node_tf}, in this paper we limit our attention to encoder-only Transformers. Encoder-only Transformers can be used for tasks such as sequence generation and sequence classification. However, in the more complicated case of a sequence-to-sequence task, an encoder-only Transformer would have to be used as input to an additional decoder module (such as a Transformer decoder or RNN decoder) that is capable of producing sequences. \par

The Transformer encoder is a special type of residual network (ResNet) in which the residual block alternates between a multi-headed self attention (MHSA) module and a dense feed-forward network (FFN) module. A given input sequence of length $L$ is first embedded into a matrix $X_0 \in \mathbb{R}^{d \times L}$, where $d$ is the embedding dimension. Letting $N$ denote the depth of the Transformer encoder (i.e., the number of blocks or layers), the hidden states $\{X_n\}_{n=1}^N \subset \mathbb{R}^{d \times L}$ of the Transformer encoder are computed according to the update rule
\begin{equation} \label{eq:tf_update_rule}
\begin{aligned}
Y_n &= X_{n-1} + \mathrm{MultiHeadedSelfAttention}_n(X_{n-1}), \\
X_n &= Y_n + \mathrm{FeedForwardNetwork}_n(Y_n),
\end{aligned}
\end{equation}
for $1 \leq n \leq N$. The operations $\mathrm{MultiHeadedSelfAttention}_n$ and $\mathrm{FeedForwardNetwork}_n$ are described in \cite{transformer} (for notational ease we will henceforth write $\mathrm{MHSA}$ and $\mathrm{FFN}$), and we have given them a subscript $_n$ to emphasize that they use different trainable parameters at each layer. Also, we have omitted in \eqref{eq:tf_update_rule} the possible use of dropout \cite{dropout} and/or layer normalization \cite{layer_norm}. \par

A drawback of vanilla Transformers is that the depth $N$ is a fixed hyperparameter, and hence does not depend on the length $L$ of the input sequence. From a practical computational efficiency standpoint, this is unsatisfactory. Indeed, consider the problem of document-level machine translation \cite{dl_translation}. When solving this problem using a vanilla Transformer, one has to choose $N$ large enough to accurately translate the longest document in the training set. This choice of $N$ may be much larger than is necessary for accurately translating shorter documents, and it is therefore natural to ask if somehow the depth $N$ could be varied depending on the length of the input sequence.  One can go even further by asking whether $N$ could be varied depending not only on $L$ but also on the ``difficulty'' (in some quantifiable sense) of the input sequence. Such a depth-adaptive Transformer could be faster to train than a vanilla Transformer, and also faster when evaluating on short/``easier'' input documents. \par

A second, theoretical drawback of $N$ being fixed, concerns the accuracy of the Transformer as the sequence length $L$ increases. In \cite{theoretical_limitations} it is shown that a Transformer with $N$ fixed layers cannot accurately model a specific class of problems, if $L$ is taken sufficiently large relative to $N$. A simple example of such a problem is \textbf{PARITY}: Given a binary string with $L$ bits, determine its parity (whether it has an odd or even number of 1s). This problem is nonlocal in the sense that it can only be solved correctly if all $L$ input bits are properly attended to, since flipping any bit in the input string will change the correct answer. A consequence of this nonlocality is that the problem is hard for vanilla Transformers to solve --- for fixed $N$ and a fixed number of attention heads it becomes increasingly difficult for a Transformer to correctly attend to all $L$ input bits if $L$ is taken increasingly large. Since correctly inferring parity is also relevant for human language, this shortcoming also limits the theoretical accuracy of natural language translation using Transformers, and is an exponent of the broader problem that Transformers need increasingly large $N$ and/or an increasing number of attention heads to translate increasingly large documents without diminishing the accuracy, due to the inherent nonlocality and context dependence of natural language documents. \par

Because \textbf{PARITY} is conceptually simple and exposes the drawbacks of having fixed $N$, we have decided to use it in all of this paper's experiments. We believe that \textbf{PARITY} serves as a simple test case for determining whether a depth-adaptive Transformer is able to properly use its depth-adaptivity to overcome fundamental drawbacks of having fixed $N$.

\section{A Neural Ordinary Differential Equation Transformer} \label{sec:node_tf}

In order to construct a depth-adaptive Transformer, we propose to replace the ResNet-based update rule \eqref{eq:tf_update_rule} with an N-ODE (see \eqref{eq:node_update_rule}, \eqref{eq:node_update_rule_skip} below). In practice, we will solve this N-ODE adaptively, with a fixed error tolerance hyperparameter for the accuracy with which the ODE trajectories are approximated by the discretized ODE. The resultant N-ODE Transformer is inherently depth-adaptive, in the sense that the number of time steps taken will vary based on how ``difficult'' it is to numerically integrate the N-ODE trajectory. \par

The equation that defines our N-ODE Transformer is chosen to be the following continuous analogue of \eqref{eq:tf_update_rule}:
\begin{equation} \label{eq:node_update_rule}
\begin{aligned}
X'(t) &= \mathrm{FFN}_t(\mathrm{MHSA}_t(X(t))), \quad 0 \leq t \leq T_F.
\end{aligned}
\end{equation}
Here $T_F$ is a fixed final time, and the dependence of $\mathrm{FFN}_t$ and $\mathrm{MHSA}_t$ on the time $t$ is described below. Motivated by the fact that \eqref{eq:tf_update_rule} describes the composition of two residual blocks, we also consider a variant of \eqref{eq:node_update_rule} in which there is a skip connection after the MHSA:
\begin{equation} \label{eq:node_update_rule_skip}
\begin{aligned}
X'(t) &= \mathrm{FFN}_t(X(t) + \mathrm{MHSA}_t(X(t))), \quad 0 \leq t \leq T_F.
\end{aligned}
\end{equation}
Although it may be possible to add dropout and/or layer normalization to \eqref{eq:node_update_rule}, \eqref{eq:node_update_rule_skip}, it does not seem straightforward what the consequences of doing so will be in the N-ODE context. We therefore do not consider dropout or layer normalization in this paper, although this could be investigated in future work. \par

The modules $\mathrm{FFN}_t$ and $\mathrm{MHSA}_t$ are parameterized with respect to $t$ by taking their time-independent vanilla counterparts and converting any trainable affine layer into a time-dependent concatenation layer, as is done in \cite{ffjord}. Explicitly, any trainable affine layer of the form
\begin{equation} \label{eq:time_ind_linear}
\begin{aligned}
\mathbb{R}^{d_1} \ni x \mapstochar \rightarrow Ax + b \in \mathbb{R}^{d_2}
\end{aligned}
\end{equation}
is converted into its time-dependent concatenation counterpart
\begin{equation} \label{eq:time_dep_linear}
\begin{aligned}
\mathbb{R}^{d_1} \ni x \mapstochar \rightarrow Ax + b + ct \in \mathbb{R}^{d_2},
\end{aligned}
\end{equation}
where $A \in \mathbb{R}^{d_2 \times d_1}$ and $b, c \in \mathbb{R}^{d_2}$ denote trainable parameters. We want to emphasize that in the vanilla Transformer \eqref{eq:tf_update_rule}, the modules $\mathrm{MHSA}_n$ and $\mathrm{FFN}_n$ have different trainable parameters for each $n$, whereas for the N-ODE Transformer \eqref{eq:node_update_rule} and \eqref{eq:node_update_rule_skip}, the modules $\mathrm{FFN}_t$ and $\mathrm{MHSA}_t$ reuse the same trainable parameters over all $t$ with the $ct$ term in \eqref{eq:time_dep_linear} encoding a dependence of the layer output on time.

We can (and will) chain together multiple N-ODE Transformer blocks, with each block having its own set of trainable parameters and time $t$ going from $0$ to $T_F$. When we do this, letting $N$ denote the total number of N-ODE Transformer blocks being chained together, the full set of equations that describes our N-ODE Transformer is given by the sequence of initial value problems
\begin{equation*}
\begin{aligned}
X_1(0) &= X_0, \\
X_i(0) &= X_{i-1}(T_F),	\quad && 2 \leq i \leq N, \\
X'_i(t) &= \mathrm{FFN}_{i,t}(\alpha X_i(t) + \mathrm{MHSA}_{i,t}(X_i(t))), \quad 0 \leq t \leq T_F, && 1 \leq i \leq N. \\
\end{aligned}
\end{equation*}
Here, $X_0 \in \mathbb{R}^{d \times L}$ is the embedding of the input sequence, and $\alpha \in \{0, 1\}$ determines whether we are choosing to add a skip connection after the MHSA. There are a total of $2N$ time-dependent modules ($\{\mathrm{FFN}_{i,t}\}_{n=1}^N$ and $\{\mathrm{MHSA}_{i,t}\}_{n=1}^N$), each module having its own set of trainable parameters. \par

In this paper, all MHSA modules are given $d / 2$ attention heads, and all FFN modules contain a single hidden layer which is dimension preserving. For all N-ODEs considered we take a final time of $T_F=1$. We solve all N-ODEs numerically using an adaptive fourth order Runge-Kutta method with absolute and relative tolerances of $10^{-5}$. The experiments that we consider in this paper are not memory intensive, and we have therefore chosen to backpropagate through all N-ODEs using automatic differentiation. For larger problems where memory efficiency is important, the adjoint method \cite{nodes} could also be used to backpropagate through N-ODEs. However, it is well-known in scientific computing that the adjoint method tends to be much slower than automatic differentiation because very small time steps are typically required to obtain an accurate approximation of the gradient, and hence we do not use it in this paper.

The reason we do not consider an N-ODE analogue of the Transformer decoder in this paper is due to the auto-regressive nature of the decoder during inference. At inference, the decoder tokens are decoded one at a time (and not all together as is the case in the encoder), and hence an N-ODE decoder would have to use interpolation to evaluate the trajectories of the previously decoded tokens at a given time, since the adaptive timestepping produces a different sequence of evaluation times each time an N-ODE is solved. It is possible to do this type of interpolation, but this is outside the scope of this paper. Our interest here is to investigate whether the depth-adaptive nature of our N-ODE Transformer can be beneficial in practice, and to investigate this we consider encoder-only N-ODE Transformers as a first step.

\section{Empirical Comparison of Different N-ODE Variants} \label{sec:rhs}

In Section \ref{sec:node_tf}, we introduced two variants of the N-ODE Transformer: one with a skip connection after the MHSA module \eqref{eq:node_update_rule_skip}, and one without \eqref{eq:node_update_rule}. We can further test whether it is worthwhile to add time dependency (via \eqref{eq:time_dep_linear}) to the N-ODE MHSA modules, versus keeping the MHSA modules time-independent (via \eqref{eq:time_ind_linear}). However, we will always stick to using time-dependent FFN modules, since others (\cite{nodes}, \cite{ffjord}) have had success with concatenation-based fully connected layers in the past. \par

We now empirically compare four N-ODE variants, by taking combinations of using or not using the MHSA skip connection and time-dependent MHSA. The test problem that we consider is \textbf{PARITY}. We prepend to all input binary strings a start of sequence (SOS) token. A given input string is first passed through an embedding layer, which is followed by the N-ODE Transformer. The final encoder hidden state of the SOS token is passed through a two layer dimension preserving fully connected network, whose output is then passed into a softmax classifier in order to predict the parity of the binary string. \par

\begin{figure}
	\centering
	\includegraphics{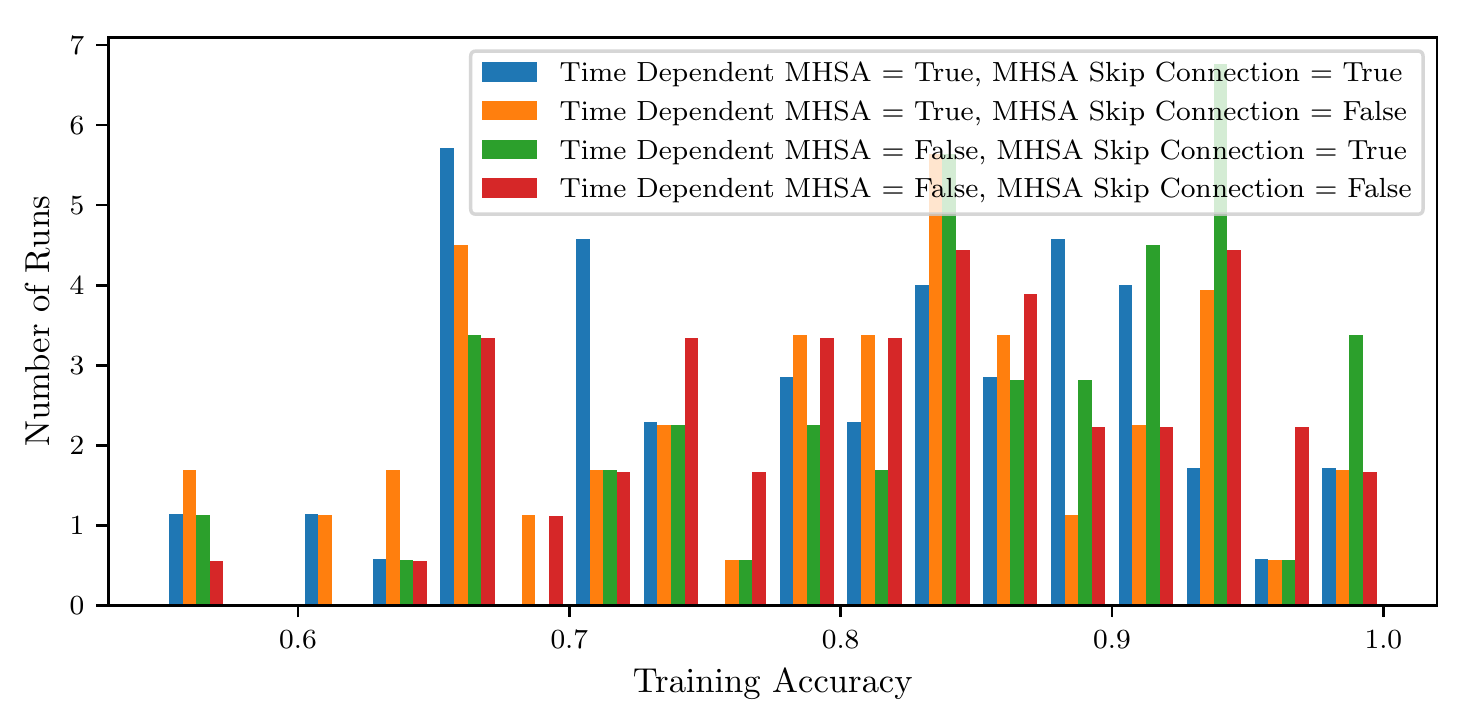}
	\caption{A histogram comparing the four N-ODE variants considered in Section \ref{sec:rhs} with
	embedding dimension $d$ = 8 and $N$ = 2 N-ODE Transformer blocks.
	Each bin in the histogram displays how many times the four variants achieved a given accuracy. Any run with an accuracy of 55\% or less is placed in the leftmost bin.}
	\label{fig:node_variants_results}
\end{figure}

For this test case\footnote{See the code at \url{https://bitbucket.org/ambaierr/node_transformer/src/master/test_tdmha_variants/}.}, we use as our training set the collection of all binary strings of length $\leq 6$ (not counting the SOS token). We use an embedding dimension of $d=8$ and $N=2$ N-ODE Transformer blocks. Each variant of the N-ODE Transformer is trained 72 times, over a variety of learning rates, in order to reduce the effect of random weight initialization and different learning rates. \par

The results of this experiment are shown in Figure \ref{fig:node_variants_results}. There do not seem to be any major accuracy differences between the four configurations. However, it appears that the red and green configurations are roughly tied for best, based on which configurations have the most runs in the two rightmost (highest accuracy) bins. Additional testing (see Appendix \ref{sec:additional_results_variants}) further suggests that the red and green configurations are roughly tied for the best. Since the red and green configurations are the ones that use time-independent MHSA, we conclude that it is best to use time-independent MHSA, and that having a skip connection after the MHSA module does not change much. \par 

In the remaining sections, all N-ODE Transformers use time-independent MHSA without a skip connection.

\section{Empirical Comparison of Vanilla and N-ODE Transformers} \label{sec:vanilla_vs_N-ODE}
We now empirically compare vanilla and N-ODE Transformers in terms of accuracy and training time\footnote{See the code at \url{https://bitbucket.org/ambaierr/node_transformer/src/master/test_node_versus_vanilla_small/}.}. As in Section \ref{sec:rhs}, we consider the \textbf{PARITY} problem, and our training set is the collection of all binary strings of length $\leq 6$ (not counting the SOS token). \par

We consider vanilla and N-ODE Transformers with embedding dimension $d \in \{4, 6, 8, 10\}$ and number of blocks $N \in \{1, 2, 3, 4\}$. By number of blocks $N$, we mean the number of Transformer blocks (or layers) in the vanilla context, and the number of N-ODE Transformer blocks in the N-ODE context (where each block performs a number of time steps). For fixed $d$, a vanilla Transformer block has roughly the same number of trainable weights as an N-ODE Transformer block, with the only difference in weight count arising from the fact that time-dependent N-ODE affine layers contain an extra time-dependent bias term (see \eqref{eq:time_dep_linear}). For each choice of $(d, N)$, the networks are trained 72 times over a variety of learning rates, in order to reduce the effect of random weight initialization and different learning rates. To summarize the results of each 72 run ensemble, we first discard the 12 worst (in terms of accuracy) runs in order to reduce the effect of outliers. For each of the remaining 60 runs, we then consider the highest achieved training accuracy of the run, and the time taken to achieve this accuracy. These quantities are averaged over the 60 runs, and as a result we have for each $(d, N)$ an \textit{(average) training accuracy} and \textit{(average) training time}. The results of this experiment are shown in Table \ref{fig:test_node_versus_vanilla_small}. \par

We see that on average, the N-ODE Transformer is 2-10 times slower to train than the vanilla Transformer. Additionally, the vanilla Transformer is consistently more accurate than the N-ODE Transformer. Indeed, for all $d \geq 8$ the vanilla Transformer achieves an accuracy of 98\% or higher, while the N-ODE Transformer is never able to reach an accuracy of 90\%. We are left to conclude that, at least for this test problem, N-ODE Transformers are appreciably worse than vanilla Transformers. \par 

\begin{table}
	\centering
	\includegraphics{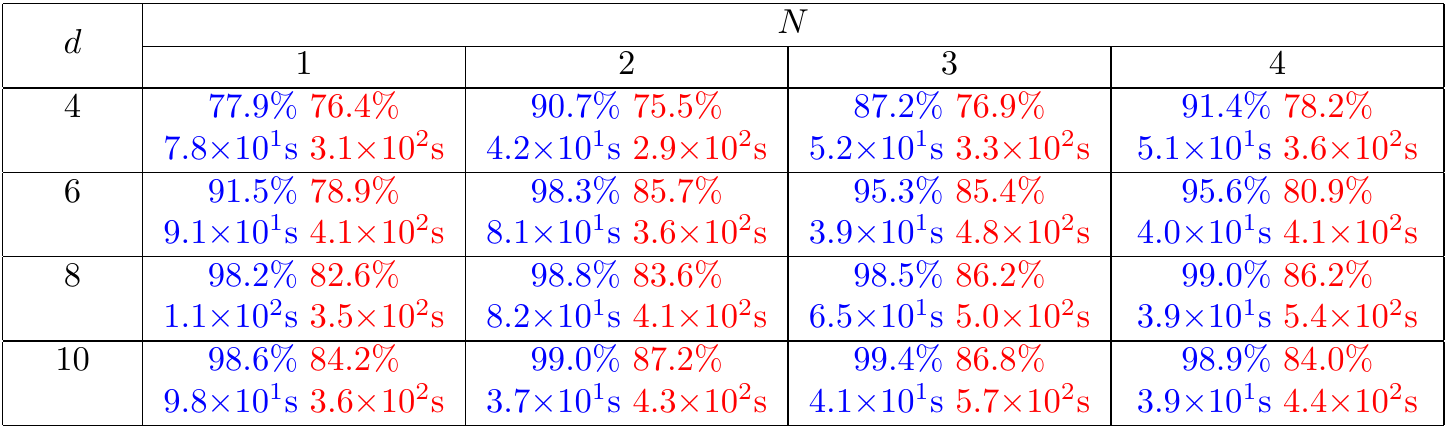}
	\caption{Results of the experiment described in Section \ref{sec:vanilla_vs_N-ODE}, shown in \textcolor{blue}{blue} for vanilla Transformers and in \textcolor{red}{red} for N-ODE Transformers. For each $(d, N)$, we show the (average) training accuracy and (average) training time. The total number of trainable weights for a given value of $(d, N)$ is about the same for the vanilla and N-ODE Transformers.
	}
	\label{fig:test_node_versus_vanilla_small}
\end{table}

Additional experiments where we have trained on longer binary strings are discussed in Appendix \ref{sec:additional_results_comparing}. Much like our findings above, these experiments suggest that N-ODE Transformers perform significantly worse than vanilla Transformers, in terms of both accuracy and training time.
More importantly, the results in Appendix \ref{sec:additional_results_comparing} illustrate that the vanilla Transformer
indeed shows quickly decreasing accuracy as the length of the input string for the \textbf{PARITY} problem increases, for example,
if one considers fixed $N=4$ and $d=10$, consistent with the theoretical limitations demonstrated in \cite{theoretical_limitations}.
However, the results in Appendix \ref{sec:additional_results_comparing} also show that the type of adaptivity provided
by the N-ODE Transformer, with the number of time steps taken in each N-ODE Transfomer block dependent on the estimated error in the ODE trajectory integration, does not help in overcoming the \textbf{PARITY} locality challenge. We will discuss in Section
\ref{sec:disc} why this is the case.

\section{Regularizing the N-ODE Trajectories} \label{sec:reg}

The findings of Section \ref{sec:vanilla_vs_N-ODE} clearly suggest that our proposed N-ODE Transformers are less accurate and slower to train than vanilla Transformers, for approximately the same number of weights. In this section, we use the regularization method of \cite{ot_flow} in an attempt to improve the performance of N-ODE Transformers. The regularization method of \cite{ot_flow} penalizes the arclength of N-ODE trajectories, and hence results in a trained network that has shorter N-ODE trajectories. The hope is that shorter N-ODE trajectories will require fewer time steps to numerically integrate, thus yielding a faster N-ODE Transformer. \par

In the N-ODE Transformer context, we implement the regularization as follows. Let $\lambda \geq 0$ be a user-chosen regularization coefficient. For each input string $\mathcal{S}$ in our training set (or more generally in our current mini-batch), and for each N-ODE Transformer block $\mathcal{B}$ in our neural network, we add to the loss the regularization term
\begin{equation} \label{eq:reg}
\frac{\lambda}{2 L} \int_0^{T_F} \mathrm{d}t \Vert X'(t)\Vert_\mathrm{Frobenius}^2
\end{equation}
where $L$ is the length of $\mathcal{S}$, and $X(\cdot) : [0, T_F] \rightarrow \mathbb{R}^{d \times L}$ is the N-ODE trajectory corresponding to $\mathcal{S}$ in the block $\mathcal{B}$. Apart from the factor of $L^{-1}$, the regularzation \eqref{eq:reg} is identical to what is introduced in \cite{ot_flow}. We have chosen to include the factor of $L^{-1}$ because different input strings will have different lengths, and hence it makes sense to normalize the regularization penalty with respect to the string length, since $\Vert X'(t)\Vert_\mathrm{Frobenius}^2$ is expected to depend linearly on $L$. \par

To empirically investigate\footnote{See the code at \url{https://bitbucket.org/ambaierr/node_transformer/src/master/test_regularization/}.} the regularization \eqref{eq:reg}, we again consider the \textbf{PARITY} problem using strings of length $\leq 6$ (not counting the SOS token). We now fix our choice of network size to be $(d, N)=(8, 2)$, and we ask what the effect of varying the regularization coefficient $\lambda$ is. There is no obvious a priori choice for $\lambda$, and we therefore experiment on the large range of $\lambda \in \{4^{(2 - i)}\}_{i=1}^{15}$. As was done in Section \ref{sec:vanilla_vs_N-ODE}, for each choice of $\lambda$ we train an ensemble of 72 runs, and only consider the best 60 of these runs in order to compute the (average) training accuracy and (average) training time. \par

\begin{figure}
	\centering
	\includegraphics{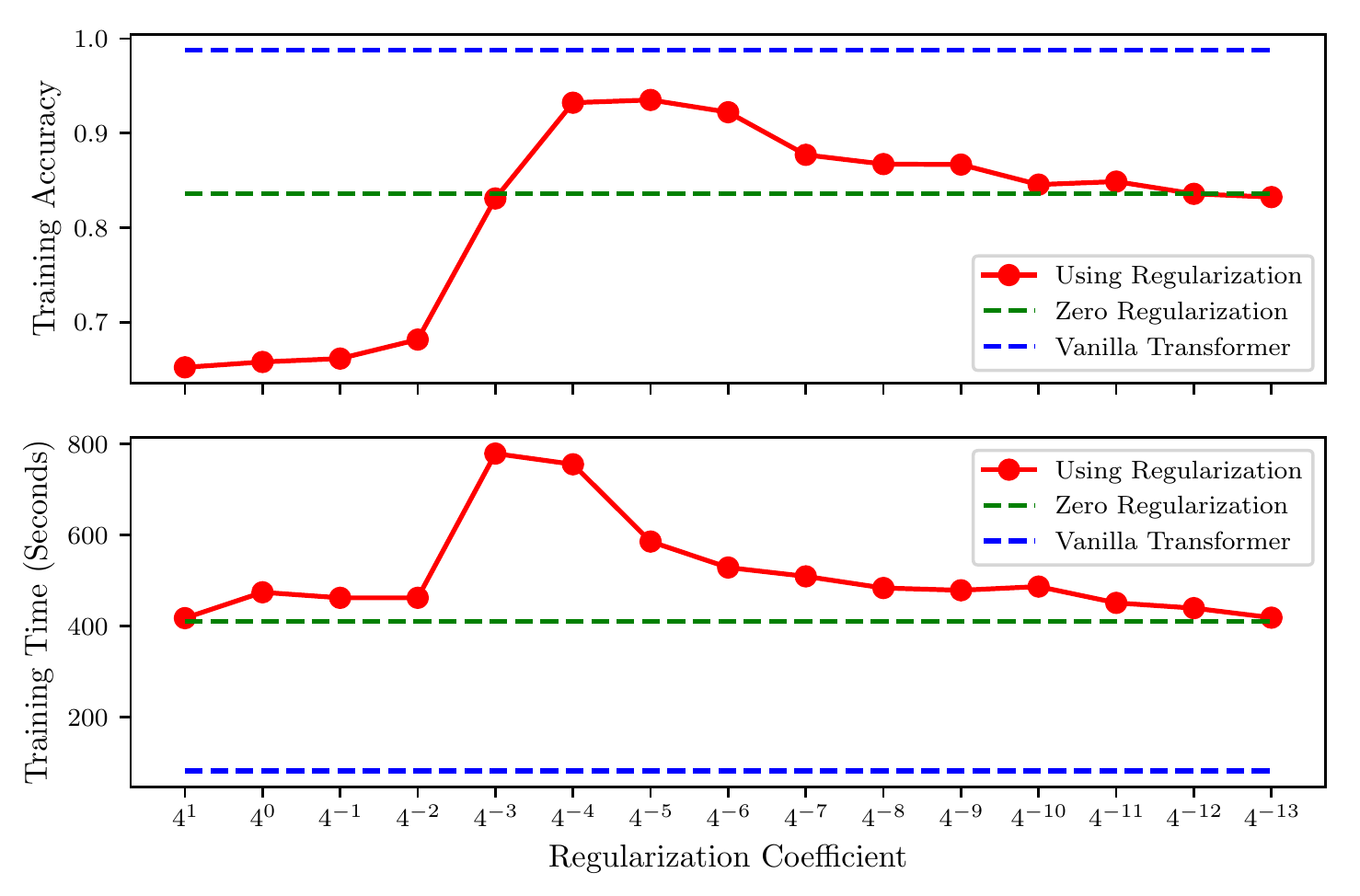}
	\caption{Results for the experiment described in Section \ref{sec:reg}. Data for the zero regularization and vanilla Transformer case is obtained from the experimental data showcased in Figure \ref{fig:test_node_versus_vanilla_small}.}
	\label{fig:reg_small}
\end{figure}

The results of this experiment are shown in Figure \ref{fig:reg_small}. Counterintuitively, we see that adding the regularization only increases the training time of the N-ODE. Fortunately, for the moderate values of $\lambda \in \{4^{-4}, 4^{-5}, 4^{-6}\}$ (which appear to be a ``sweet spot'' for the choice of $\lambda$), the increase in training time comes with an appreciable increase in accuracy. Still, the N-ODE Transformer never reaches the accuracy achieved by the vanilla Transformer, and in all cases the N-ODE Transformer takes at least 5 times longer to train than the vanilla Transformer. \par

Why do values of $\lambda \in \{4^{-4}, 4^{-5}, 4^{-6}\}$ lead to higher accuracies but also higher training times? In the case of no regularization ($\lambda=0$) the learned N-ODE trajectories quickly become long and complicated. We hypothesize that the optimization landscape corresponding to such N-ODE trajectories is rough and contains many sharp local minima. In this case the optimizer quickly reaches a maximum attainable accuracy, and it cannot improve the accuracy any further. By taking values of $\lambda$ in the ``sweet spot'' of $\{4^{-4}, 4^{-5}, 4^{-6}\}$, the resultant optimization landscape is relatively smooth, and a higher maximum accuracy can be achieved at the cost of taking longer to reach it. \par

In Appendix \ref{sec:additional_regularization_results}, we have repeated the experiment outlined above on longer binary strings. Unfortunately, in this case we find that for no values of $\lambda$ does regularization substantially improve the accuracy or training time. These findings lead us to conclude that while the regularization \eqref{eq:reg} may be useful for small problems, it is likely detrimental for larger problems. Indeed, for larger problems it may be necessary to allow for long and complicated N-ODE trajectories. In this case, other regularization approaches such as \cite{node_reg} may be worth investigating.

\section{Discussion and Future Work} \label{sec:disc}

In this paper, we have motivated and introduced an N-ODE variant of the Transformer. We empirically found that, at least on the \textbf{PARITY} problem, the N-ODE Transformer is inferior to the vanilla Transformer in terms of accuracy and training time. 
We also found that the type of depth-adaptivity the N-ODE Transformer provides does not help to address the known
difficulty of the Transformer in handling highly nonlocal problems such as \textbf{PARITY}, even though increasing the
number of layers in the vanilla Transformer should address this problem. We also investigated an N-ODE regularization scheme, and found that it improves the accuracy of the N-ODE Transformer in the case of a sufficiently small training set, but appears to not be beneficial on larger training sets. \par

In order to make an attempt at explaining these observations, it is useful to consider the relation between the discretized N-ODE Transformer and the N-ODE it discretizes. Suppose that we write our N-ODE with parameters $\theta$ in the abstract form
\begin{equation} \label{eq:node_abstract}
X'(t) = F(X(t), t; \theta), \quad 0 \leq t \leq T_F.
\end{equation}
For simplicity, let us discretize \eqref{eq:node_abstract} using the forward Euler method with $N$ time steps. This results in the discrete update rule
\begin{equation} \label{eq:node_abstract_discrete}
X_{n+1} = X_n + T_F \frac{1}{N} F(X_n, T_F \frac{n}{N}; \theta), \quad 0 \leq n \leq N-1.
\end{equation}
A first question that arises is whether it is reasonable to interpret \eqref{eq:node_abstract_discrete} as an $N$ layer ResNet.
When we use the standard concatenation approach of (\ref{eq:time_dep_linear}) 
(as in \cite{nodes,ffjord}) to define the right-hand side function of N-ODE (\ref{eq:node_abstract}),
each time step or ``layer'' in \eqref{eq:node_abstract_discrete} uses the same parameters $\theta$, and therefore \eqref{eq:node_abstract_discrete} is at best a time-dependent enhancement of a ResNet with weight sharing. 
So it is not obvious whether \eqref{eq:node_abstract_discrete} with time-dependent parameterizations as in (\ref{eq:time_dep_linear})  can be viewed as being comparable to an $N$-layer ResNet.

The theoretical results in \cite{theoretical_limitations} predict that increasing the number of layers in ResNets 
will address the accuracy issues the vanilla Transformer faces when dealing with highly nonlocal problems
such as \textbf{PARITY}. At first sight, one may think that the depth-adaptivity of the N-ODE Transformer (within
each block) could similarly help information to propagate over longer distances due to the additional time step
layers, but unfortunately it appears that due to the effective weight sharing the additional time step layers
cannot properly exploit this opportunity to increase connections over longer distances in the sequence.
So in hindsight, this is not an unexpected finding. On the contrary, the reason vanilla Transformers can mitigate
this issue by adding more layers, appears to be connected directly to the fact that every additional layer
in a vanilla Transformer brings with it an additional set of trainable weights.
Another way to see why adding more time step layers in N-ODE Transformer blocks may not help to improve
the expressivity of the network goes as follows. As $N \rightarrow \infty$, the residuals
$T_F \frac{1}{N} F(X_n, T_F \frac{n}{N}; \theta)$ in \eqref{eq:node_abstract_discrete} converge to zero.
Hence for $N$ sufficiently large the impact of a given ``layer'' in \eqref{eq:node_abstract_discrete} on the
final state $X_N$ is negligible, in particular if one realizes (again) that the weights are shared between
all the layers.
Next, a plausible explanation for why our proposed N-ODE Transformer has worse accuracy and training times than
the vanilla Transformer is that the N-ODE Transformer blocks with multiple adaptive time steps may lead to
harder optimization problems. \par

More generally, it is not well documented or understood in which context the type of adaptivity that is advertised for N-ODEs 
\cite{nodes,ffjord} can actually be useful for deep learning problems. First, N-ODE adaptivity is driven by
estimating the error by which the N-ODE discretization approximates the continuous trajectories of the N-ODE,
but it is by no means clear to which extent reducing this trajectory error by adaptivity contributes to reducing the
training or prediction error of the network. The real goal should be to reduce the prediction error, and
methods should be developed where the adaptivity is driven directly by reduction in the prediction error.
Similarly, \cite{nodes} advocates for solving N-ODEs using high-order ODE discretizations, but here too it is
not clear how the use of high-order methods would lead to more accurate predictions. High-order methods
are beneficial for accurately approximating sufficiently smooth trajectories with reduced cost, but again the link between
improved accuracy of trajectories and reduced prediction error for the neural network loss is far
from clear, except perhaps in very specific applications such as the prediction of time series trajectories
that can be modelled by N-ODEs.
\par

The investigation and analysis that we have conducted in this paper on N-ODE versions of Transformer networks is quite non-exhaustive. To the best of our knowledge, there have not been attempts to combine N-ODEs and Transformers in the past. Therefore, there is much more work that could be done towards designing and understanding N-ODE Transformers. Examples of interesting avenues for future work include:

\begin{itemize}
	\item \textbf{Investigating other ODE analogues of the Transformer}. The Transformer update rule \eqref{eq:tf_update_rule} can be compactly written as
	\begin{equation*}
		X_n = X_{n-1} + \mathrm{MHSA}_n(X_{n-1}) +  \mathrm{FFN}_n(X_{n-1} + \mathrm{MHSA}_n(X_{n-1})).
	\end{equation*}
	This bears resemblance to a forward Euler discretization of the ODE
	\begin{equation} \label{eq:other_analogue}
	X'(t) = \mathrm{MHSA}_t(X(t)) + \mathrm{FFN}_t(X(t) + \mathrm{MHSA}_t(X(t))).
	\end{equation}
	An N-ODE of the form \eqref{eq:other_analogue} may be worth investigating as an alternative to \eqref{eq:node_update_rule} or \eqref{eq:node_update_rule_skip}. This N-ODE Transformer variant may
	improve accuracy and training time, possibly to a level comparable to the vanilla Transformer, since
        $N$ of these N-ODE Transformer blocks with one time step per block would be very close to a vanilla
        Transformer with $N$ blocks. 
	\item \textbf{Using layer normalization and dropout in the N-ODE}. Layer normalization can be viewed as a type of regularization, and hence may be useful in the N-ODE context. In addition, if the N-ODE Transformer were to be used on large problems where overfitting needs to be dealt with, dropout could be useful.
	\item \textbf{Investigating other regularization schemes}. The regularization that we employed in Section \ref{sec:reg} is a special case of the regularization scheme proposed in \cite{node_reg} (in the notation of \cite{node_reg} it corresponds to $K=1$, where $K$ is the order of the trajectory derivative that is being penalized). It may be worthwhile to consider the more general regularization scheme proposed in \cite{node_reg} with $K > 1$.
	\item \textbf{Implementing an N-ODE decoder}. It may be better to use depth-adaptivity in the decoder, and not in the encoder (for instance, this approach is taken in \cite{depth_adaptive_transformer}).
	\item \textbf{Exploring other test problems}. Our empirical tests are limited to the \textbf{PARITY} problem, which as we mentioned in Section \ref{sec:tf} is nonlocal. Perhaps the N-ODE Transformer would fare better on a semilocal problem such as (non context-aware) translation.
	\item \textbf{Investigating the theoretical limitations of N-ODE Transformers}. The N-ODE Transformer is depth-adaptive, provided we interpret depth as being the number of time steps. Empirically, the N-ODE Transformer fares worse on \textbf{PARITY} than the vanilla Transformer. Can it be shown theoretically that in spite of its depth-adaptivity, the N-ODE Transformer has the same theoretical limitations (see \cite{theoretical_limitations}) as the vanilla Transformer?
\end{itemize}

\section*{Acknowledgement}
We thank Reza Haffari for interesting discussions and ideas.

\bibliographystyle{unsrt}
\bibliography{references}

\newpage

\appendix
\renewcommand{\thesubsection}{\Alph{subsection}}
\subsection{Additional Results Comparing N-ODE Variants} \label{sec:additional_results_variants}
\begin{figure}
	\centering
	\includegraphics{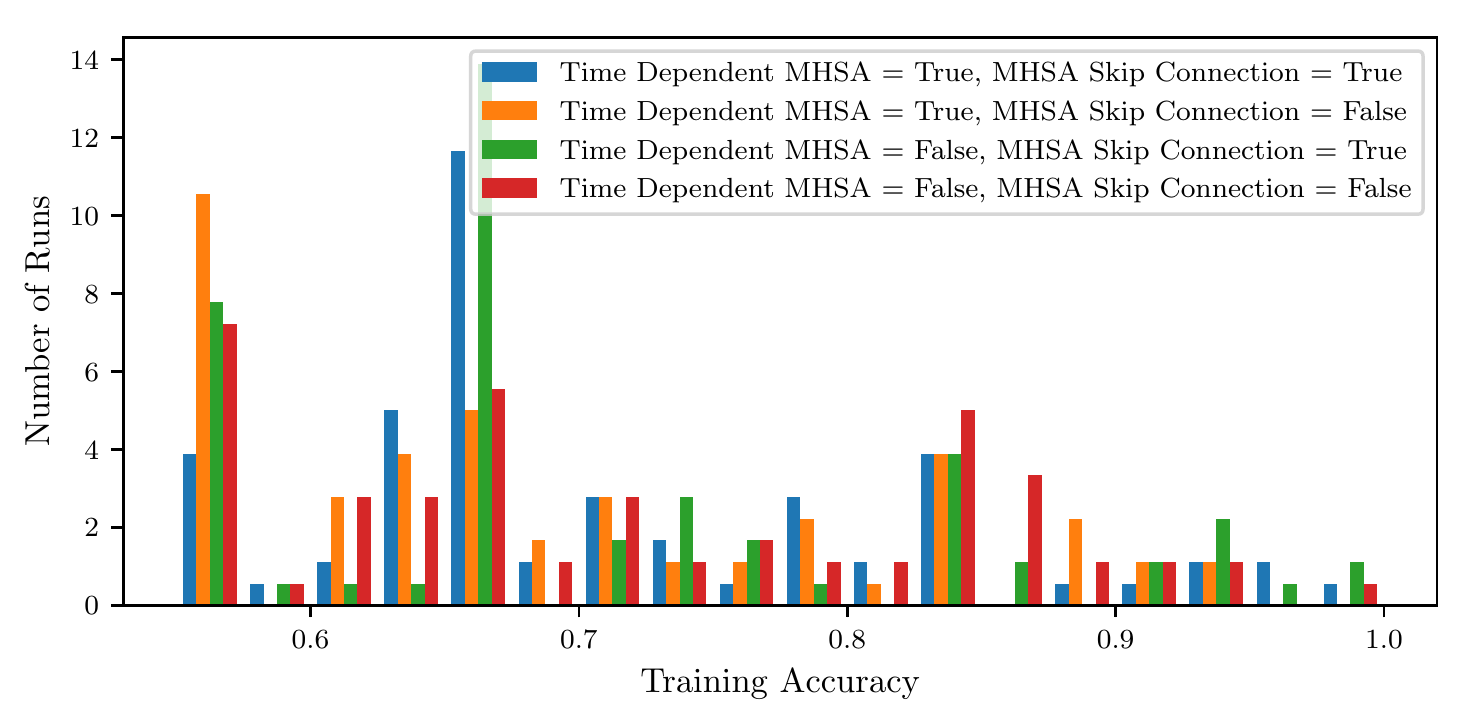}
	\caption{Results as in Fig.\ \ref{fig:node_variants_results} for embedding dimension $d$ = 4 and $N$ = 1 N-ODE Transformer blocks.}
	\label{fig:node_variants_1}
\end{figure}

\begin{figure}
	\centering
	\includegraphics{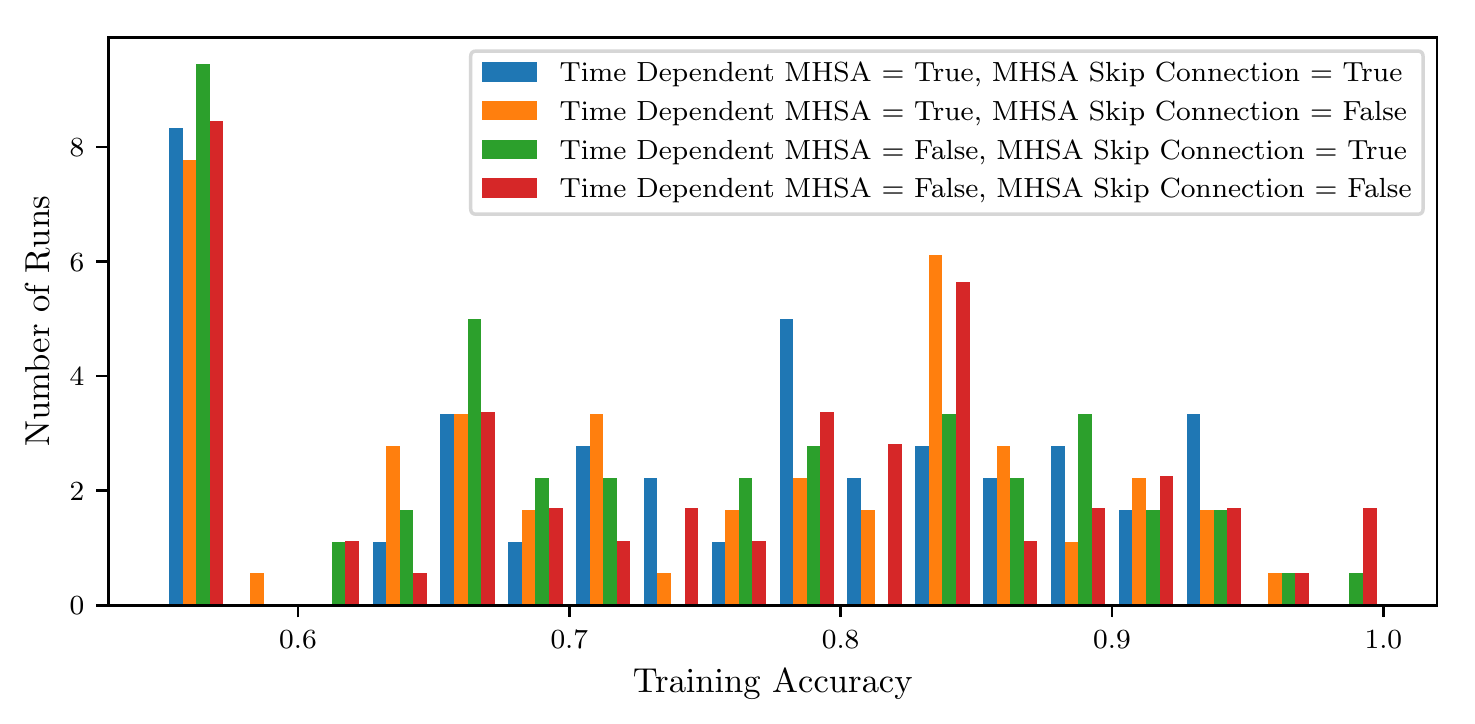}
	\caption{Results as in Fig.\ \ref{fig:node_variants_results} for embedding dimension $d$ = 4 and $N$ = 2 N-ODE Transformer blocks.}
	\label{fig:node_variants_2}
\end{figure}

\begin{figure}
	\centering
	\includegraphics{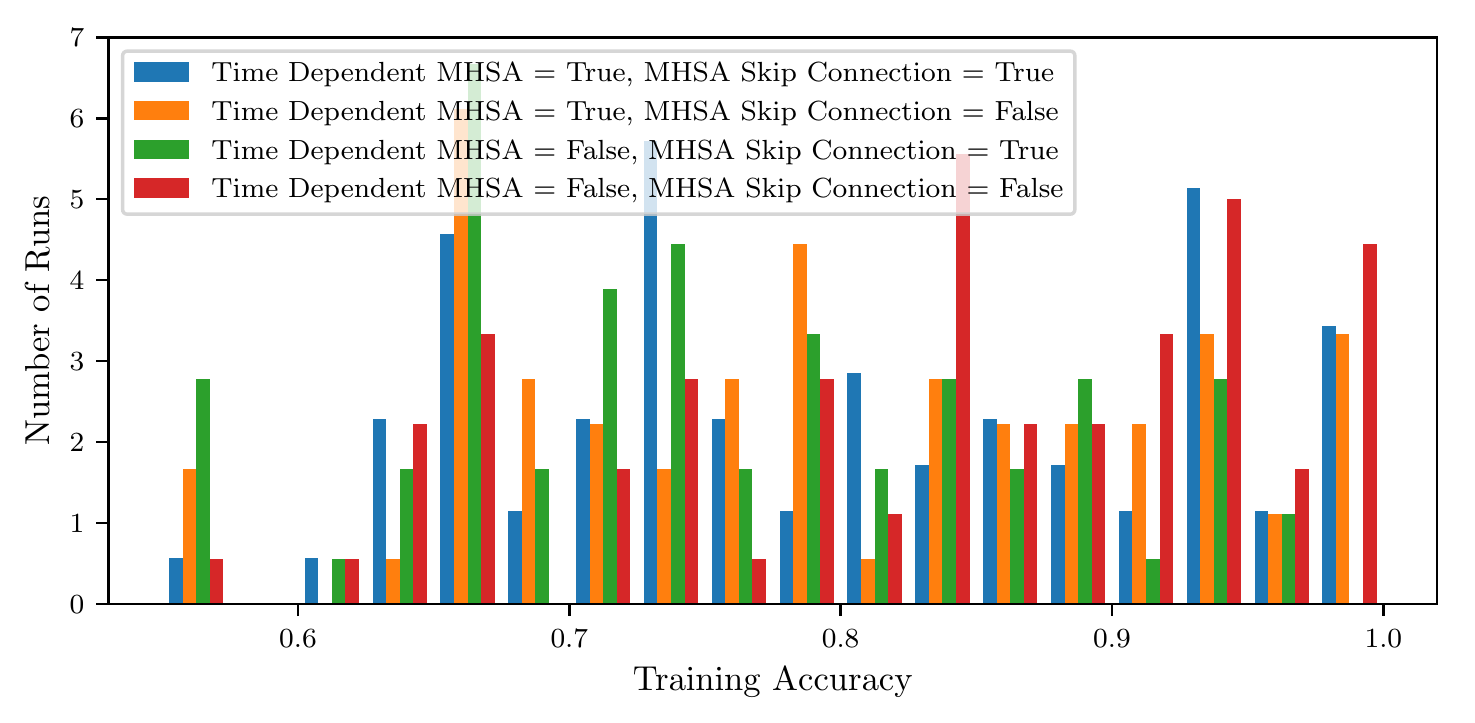}
	\caption{Results as in Fig.\ \ref{fig:node_variants_results} for embedding dimension $d$ = 8 and $N$ = 1 N-ODE Transformer blocks.}
	\label{fig:node_variants_3}
\end{figure}

In this Appendix section, we have repeated the experiment of Section \ref{sec:rhs} three additional times, each time with a different choice of $(d, N)$. The results are shown in Figures \ref{fig:node_variants_1}, \ref{fig:node_variants_2}, \ref{fig:node_variants_3}. In all cases, it is either the red or green configuration that gets the most runs in the rightmost (highest accuracy) bin. This further confirms what we found in Section \ref{sec:rhs}.

\newpage

\subsection{Additional Results Comparing Vanilla and N-ODE Transformers} \label{sec:additional_results_comparing}

In this Appendix section, we have repeated the test case described in Section \ref{sec:vanilla_vs_N-ODE}, but we have increased the length of the binary strings in our training set. In Tables \ref{fig:test_node_versus_vanilla_medium} and \ref{fig:test_node_versus_vanilla_big} are our experimental findings with binary strings of length $\leq 8$ and $\leq 10$. \par

We find that while the N-ODE Transformer remains slower to train than the vanilla Transformer, the two do have training times that lie within the same order of magnitude. This is in contrast to Section \ref{sec:vanilla_vs_N-ODE}, where the vanilla Transformer was sometimes up to 10 times faster to train than the N-ODE Transformer. \par

In terms of accuracy, both Tables \ref{fig:test_node_versus_vanilla_medium} and \ref{fig:test_node_versus_vanilla_big} show that the vanilla Transformer is often more accurate than the N-ODE Transformer by an amount of 20\% or more. In fact, looking at Table \ref{fig:test_node_versus_vanilla_big}, we see that the N-ODE Transformer barely reaches an accuracy of 60\%, and concerningly its accuracy never improves as $d$ or $N$ are increased.

\begin{table}
	\centering
	\includegraphics{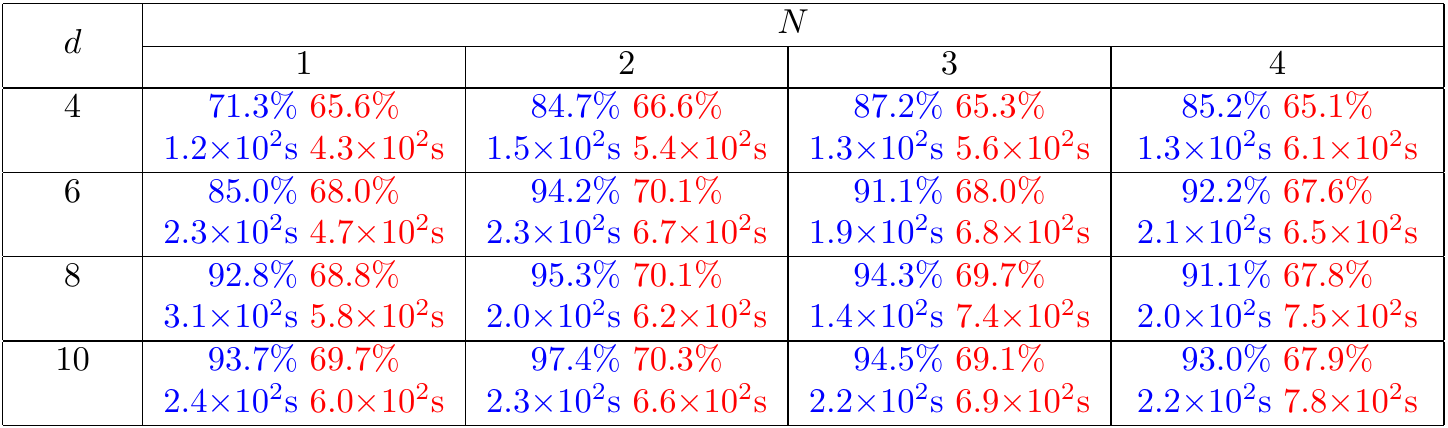}
	\caption{Results on binary strings of length $\leq 8$ (not counting the SOS token). Results are shown in \textcolor{blue}{blue} for vanilla Transformers and in \textcolor{red}{red} for N-ODE Transformers.}
	\label{fig:test_node_versus_vanilla_medium}
\end{table}

\begin{table}
	\centering
	\includegraphics{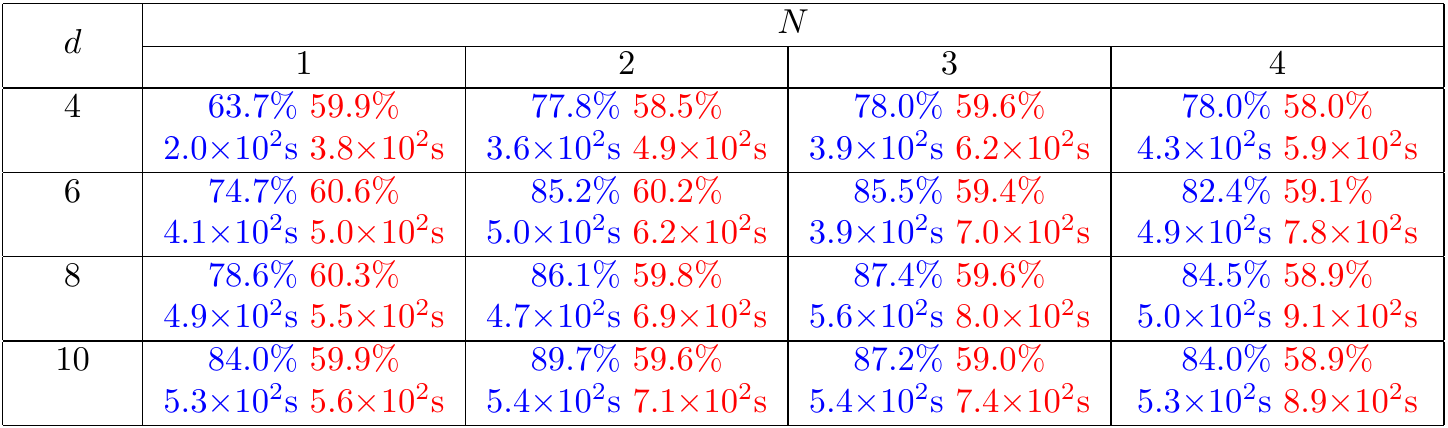}
	\caption{Results on binary strings of length $\leq 10$ (not counting the SOS token). Results are shown in \textcolor{blue}{blue} for vanilla Transformers and in \textcolor{red}{red} for N-ODE Transformers.}
	\label{fig:test_node_versus_vanilla_big}
\end{table}

\newpage 
\subsection{Additional Regularization Results} \label{sec:additional_regularization_results}
\begin{figure}
	\centering
	\includegraphics{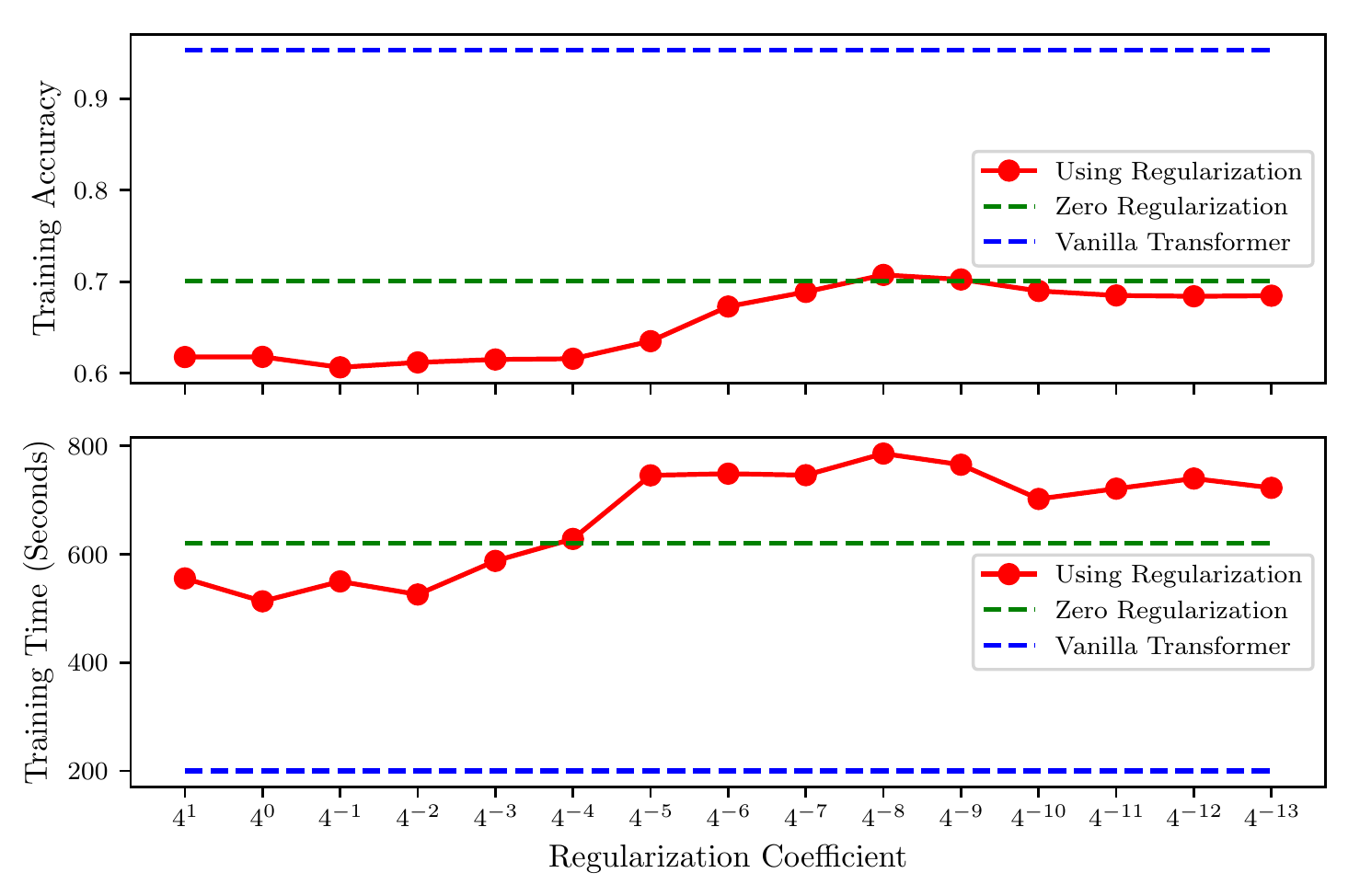}
	\caption{Results on binary strings of length $\leq 8$ (not counting the SOS token).}
	\label{fig:reg_medium}
\end{figure}

\begin{figure}
	\centering
	\includegraphics{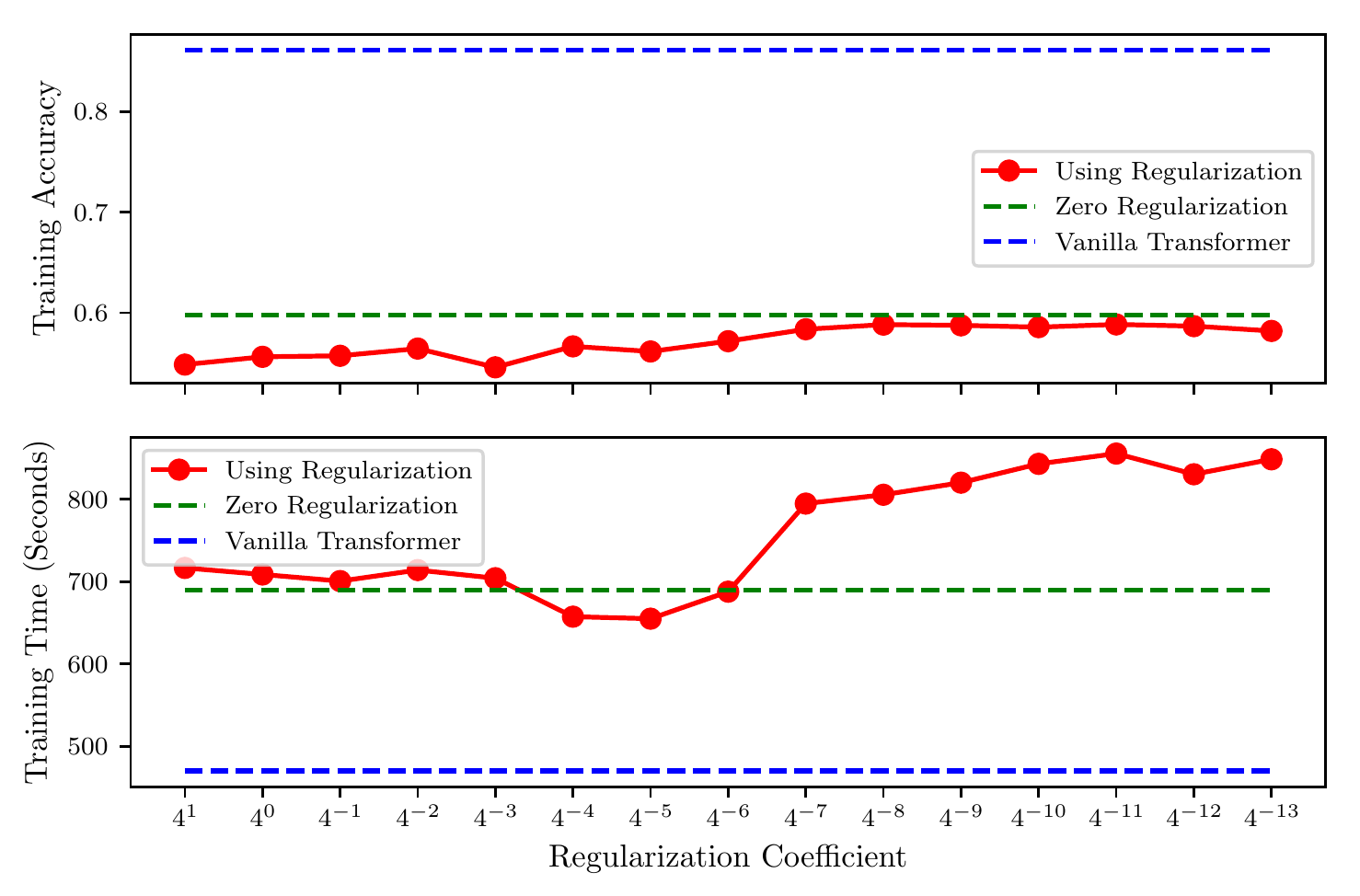}
	\caption{Results on binary strings of length $\leq 10$ (not counting the SOS token).}
	\label{fig:reg_big}
\end{figure}

In this Appendix section, we have repeated the test case described in Section \ref{sec:reg}, but we have increased the length of the binary strings in our training set. In Figures \ref{fig:reg_medium} and \ref{fig:reg_big} are our experimental findings with binary strings of length $\leq 8$ and $\leq 10$. \par

\end{document}